\PassOptionsToPackage{dvipsnames,table}{xcolor}
\documentclass[sigconf=true, nonacm=true, review=false, anonymous = false]{acmart}
\usepackage{booktabs}
\usepackage{amsmath}
\usepackage{xspace}
\usepackage{xcolor}
\usepackage{dsfont}
\usepackage{rotating}
\usepackage{makecell}

\usepackage{amsthm}
\usepackage{subcaption}
\renewcommand{\epsilon}{\varepsilon}

\usepackage{mathtools}

\begin{document}

\title{Using Affine Combinations of BBOB Problems for Performance Assessment}

\author{Diederick Vermetten}
\orcid{0000-0003-3040-7162}
\affiliation{
  \institution{Leiden Institute for Advanced Computer Science}
  \city{Leiden}
  \country{The Netherlands}
}
\email{d.l.vermetten@liacs.leidenuniv.nl}

\author{Furong Ye}
\orcid{0000-0002-8707-4189}
\affiliation{
  \institution{Leiden Institute for Advanced Computer Science}
  \city{Leiden}
  \country{The Netherlands}
}
\email{f.ye@liacs.leidenuniv.nl}

\author{Carola Doerr}
\orcid{0000-0002-4981-3227}
\affiliation{
  \institution{Sorbonne Universit\'e, CNRS, LIP6}
  \city{Paris}
  \country{France}}
\email{Carola.Doerr@lip6.fr}

\begin{abstract} 
Benchmarking plays a major role in the development and analysis of optimization algorithms. 
As such, the way in which the used benchmark problems are defined significantly affects the insights that can be gained from any given benchmark study.
One way to easily extend the range of available benchmark functions is through affine combinations between pairs of functions. 
From the perspective of landscape analysis, these function combinations smoothly transition between the two base functions.

In this work, we show how these affine function combinations can be used to analyze the behavior of optimization algorithms. In particular, we highlight that by varying the weighting between the combined problems, we can gain insights into the effects of added global structure on the performance of optimization algorithms. By analyzing performance trajectories on more function combinations, we also show that aspects such as the scaling of objective functions and placement of the optimum can greatly impact how these results are interpreted. 
\end{abstract}


\maketitle

\section{Introduction}
Benchmarking is a key aspect in the development of optimization algorithms. Not only are benchmark problems used to compare the effectiveness of different optimizers with regard to a standardized set of problems, the analysis of algorithm behavior on these problems is often used to gain insight into the characteristics of the algorithm. Because of this, the design of benchmark problems has a major impact on the field of optimization as a whole~\cite{benchmarkingOpt}.

One of the most common benchmark suites in single-objective, continuous, noiseless optimization is fittingly called Black Box Optimization Benchmark (BBOB)~\cite{bbobfunctions}. This suite is part of the COCO framework~\cite{hansen2020coco}, which has seen significant adoption in the last decade. This suite consists of 24 problems, each defined to represent a set of global landscape properties. For each of these problems, many different instances can be created through a set of transformations, allowing researchers to test different invariances of their algorithm. Because of its popularity, studies into the specifics of the BBOB suite are numerous~\cite{munoz2015algorithm, evostar_bbob_instance, ela2_munoz2022}. 

One particularly popular method to investigate continuous optimization problems is Exploratory Landscape Analysis (ELA)~\cite{mersmann2011exploratory}. This technique aims to characterize the low-level landscape properties through a large set of features. Applying this to the BBOB suite shows that instances of the 24 functions generally group together, with separation between functions being relatively robust~\cite{renau2021towards}. This observation raised the question of how the spaces between problems could be explored. 

In a recent study, affine combinations between pairs of BBOB problems were proposed and analyzed using ELA~\cite{affinebbob}. The resulting analysis shows that varying the weight of these combinations has a relatively smooth impact on the landscape features. As such, these new functions could potentially be used to study the transition between different landscapes, which opens up a more in-depth analysis of the relation between landscapes and algorithm behavior. 

To investigate to what extent the affine function combinations can be used to study algorithmic behavior, we perform a benchmarking study through which we investigate the effect of the affine combinations on the performance of five numerical black-box optimization algorithms. We make use of function combinations which include a sphere model to show the impact of added global structure on the relative ranking between algorithms. Additionally, we show that by combining functions with different global properties we don't always obtain smooth transitions in performance. We provide examples where the combination of two functions can either be significantly more challenging or slightly easier than the base functions it consists of.

\section{Related Work}

\subsection{BBOB Problem Suite}

Within continuous optimization benchmarking, one of the most popular suites of benchmarks is the BBOB family, which has been designed as part of the COCO framework. The noiseless, single-objective suite consists of 24 problems, each of which can be instantiated with a set of different transformations. These function instances aim to preserve the global function properties while varying factors such as the location of the global optimum, such that an optimizer can not directly exploit these aspects. However, the exact influence these transformations have on the low-level landscape properties is not as straightforward, which can lead to noticeable differences in algorithm behavior on different instances of the same function~\cite{evostar_bbob_instance}. 

\subsection{Affine Function Combinations}
While using function instances allows the BBOB suite to cover a wider range of problem landscapes than the raw functions alone, there are limits to the types of landscapes which can be created in this way. Recently, it has been proposed to use affine combinations between pairs of BBOB functions to generate new benchmark functions~\cite{affinebbob}. These combinations have been shown to smoothly fill the space of low-level landscape properties, as measured through a set of ELA features. These results have shown that even a relatively simple function creation procedure has the potential to give us new insights into the way function landscapes work.

\section{Experimental Setup}

\begin{figure*}[th]
    \centering
    \includegraphics[width=0.97\textwidth]{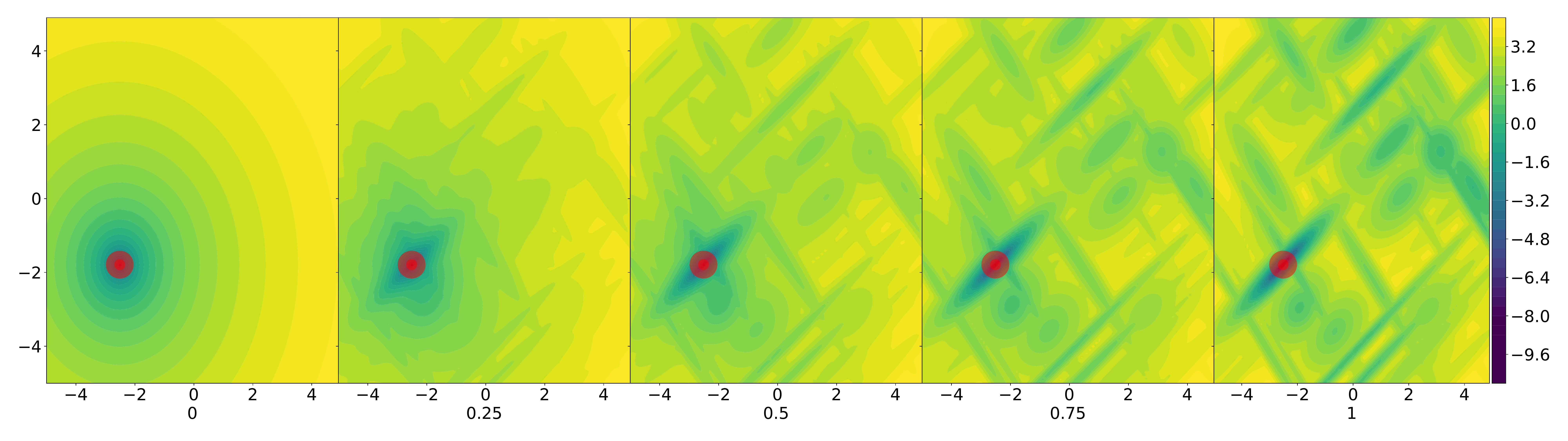}
    \caption{Evolution of the landscape (log-scaled function-values) of the affine combination between F21 ($\alpha=1$) and F1 ($\alpha=0$), instance 1 for both functions, for varying $\alpha$. The red circle highlights the location of the global optimum.}
    \label{fig:frame_f21_f1}
\end{figure*}

In this work, we make use of a slightly modified version of the affine function combinations from~\cite{affinebbob}. In particular, we define the combination between two functions from the BBOB suite as follows:
\begin{align*}
C(F_1, I_1, F_2, I_2,\alpha)(x) =& \\ 
\exp\Big( \alpha &\log\big(F_1(x) - F_1(O_1)\big) + \\ 
         (1-\alpha) &\log\big(F_2(x - O_1 + O_2) - F_2(O_2) \big)\Big) 
\end{align*}

Where $F_1$, $I_1$, $F_2$, $I_2$ are the two base functions and their instance number, as defined in BBOB~\cite{bbobfunctions}. $O_1$ and $O_2$ represent the location of the optimum of functions $F_1$ and $F_2$ respectively.  The transformation to $x$ when evaluating $F_2$ is performed to make sure the location of the optimum is at $O_1$. As opposed to the original definition, we subtract the optimal values before aggregating, so we can take a logarithmic mean between the problems. This way, we can use consistent values for $\alpha$ across problems, without having to perform the entropy-based selection performed in~\cite{affinebbob}. It has the additional benefit of ensuring the objective value of the optimal solution is always 0, so the comparison of performance across instances and across problems is simplified. In Figure~\ref{fig:frame_f21_f1}, we illustrate the change in landscape for the combination of F21 and F1, for different values of $\alpha$. 

In order to implement these function combinations, we make use of the IOHexperimenter~\cite{iohexp} framework. We access the BBOB problems, combine them together as described, and wrap them into a new problem. This enables us to use any of the built-in logging and tracking options of IOHexperimter. In particular, it allows us to store the performance data into a file-format which can be directly processed into IOHanalyzer~\cite{IOHanalyzer} for post-processing. 

For our algorithm portfolio, we make use of the Nevergrad toolbox, which provides a common interface to a wide range of optimization algorithms~\cite{nevergrad}. In this study, we benchmark the following algorithms:
\begin{itemize}
    \item Particle Swarm Optimization (PSO)~\cite{pso}
    \item Constrained Optimization BY Linear Approximation (\sloppy{Cobyla}) ~\cite{cobyla}
    \item Differential Evolution (DE) ~\cite{de}
    \item Estimation of Multivariate Normal Algorithm (EMNA)~\cite{emna}
    \item Diagonal Covariance Matrix Adaptation Evolution Strategy (dCMA-ES)~\cite{hansen2001self_adaptation_es}
\end{itemize}
For each of these algorithms, we make use of the default parameters as chosen in Nevergrad. Each run of the algorithm has a budget of $2\,000 D$, where $D$ is the dimension of the problem. We perform $5$ independent runs per instance. In the remainder of this paper, we set $I_2=1$. As such, when discussing the \emph{instance} of an affine function combination $C(F_1, I_1, F_2, I_2, \alpha)$, we are referring to $I_1$. 

\textbf{Reproducibility} To ensure reproducibility, we make all code used in the creation of this paper available in a Zenodo repository~\cite{reproducibiliyt}. This repository contains the data generation code, raw data generated, and post-processing scripts used to create the results discussed in the following sections, following the recommendations proposed in ~\cite{lopez2021reproducibility}. In addition to this, we also make available a Figshare repository containing additional figures and animations which could not be included in this paper~\cite{reproducibiliyt}.

\section{Performance Comparison for Affine Combinations With F1}
For a first set of experiments, we make use of affine combinations where we combine each function with F1: the sphere model. As can be seen in Figure~\ref{fig:frame_f21_f1}, adding a sphere model to another function creates an additional global structure that can guide the optimization toward the global optimum. As such, these kinds of combinations might allow us to investigate the influence of an added global structure on the performance of optimization algorithms. While to some extent this can already be investigated by comparing results on the function groups of the original BBOB with different levels of global structure, the affine function combinations allow for a much more fine-grained investigation. Since the landscape features of these combined functions seem to shift smoothly when varying $\alpha$, we might assume similar behavior on algorithmic performance. 

\begin{figure}
    \centering
    \includegraphics[width=0.48\textwidth]{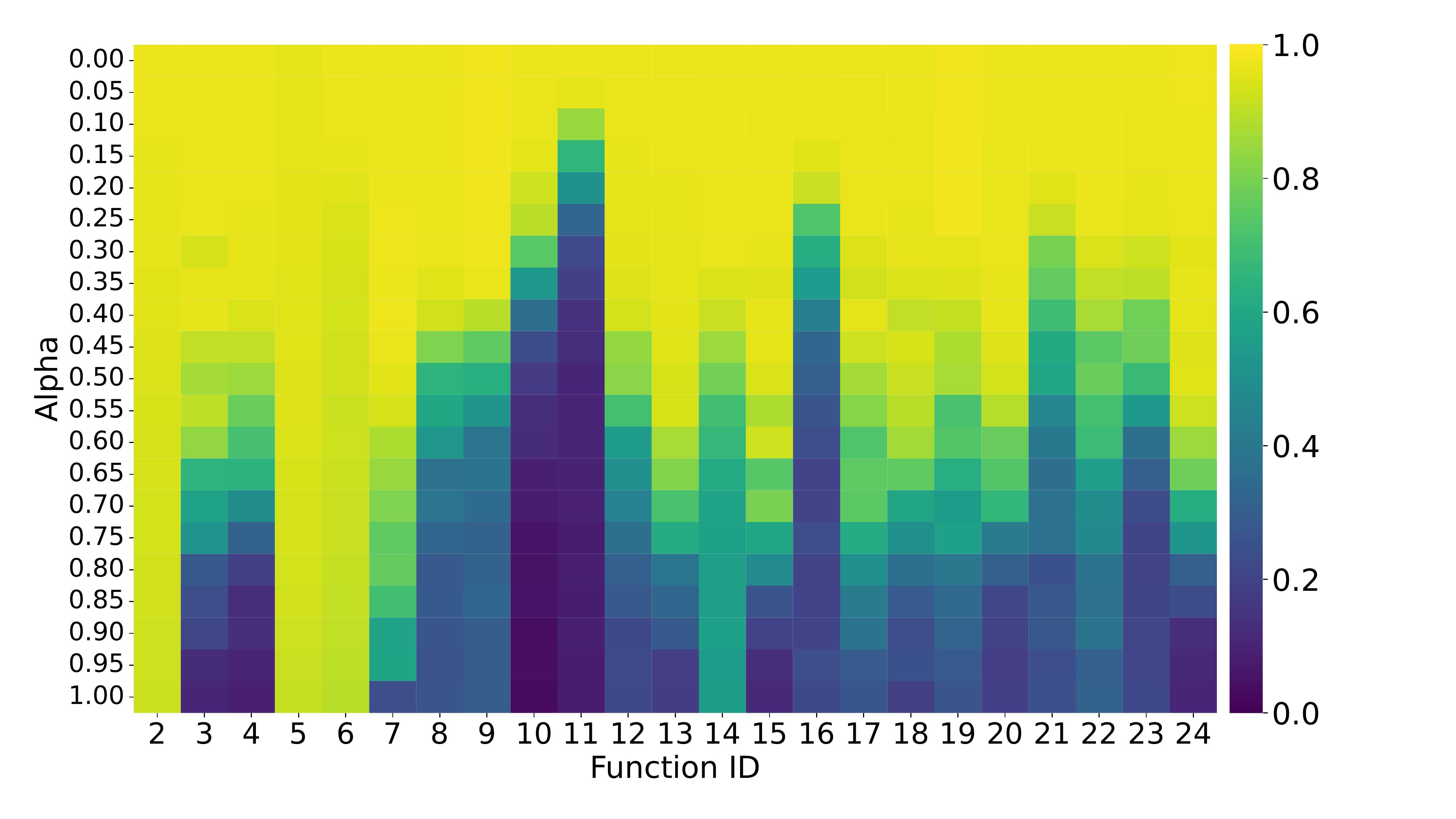}
    \caption{Normalized area under the ECDF curve of Diagonal CMA-ES for each combination of the BBOB-function (x-axis) with a sphere model, for given value of $\alpha$ (y-axis). AUC is calculated after $10\,000$ function evaluations, based on 50 runs on 10 instances. }
    \label{fig:cma_heatmap_wsphere}
\end{figure}

In Figure~\ref{fig:cma_heatmap_wsphere}, we show the performance of diagonal CMA-ES, measured as the area under the Empirical Cumulative Distribution Function (ECDF)~\cite{hansen2022anytime}, for varying function combinations and $\alpha$ values. As is widely accepted for BBOB functions, we make use of 51 targets logarithmically spaced between $10^{2}$ and $10^{-8}$ to compute the ECDF. The resulting Area Under the Curve (AUC) is normalized, so an algorithm which reaches all targets in the first evaluation would have an AUC of $1$. The top of this figure, with $\alpha = 0$, shows the performance on the sphere function, on which CMA-ES performs very well. There are however differences between the columns, since the location of the affine function combination is set to the optimum of the second function. 

In Figure~\ref{fig:cma_heatmap_wsphere}, we can see that the performance of CMA-ES does indeed seem to move smoothly between the sphere and the function with which it is combined. It is however interesting to note the differences in speed at which this transition occurs. While the final performance on e.g. functions 3 and 11 seems similar, the transition speed differs significantly. This seems to indicate that for F11, the addition of some global structure has a relatively weak influence on the challenges of this landscape from the perspective of the CMA-ES, while even small amounts of global structure significantly simplify the landscape of F3.

\begin{figure}
    \centering
    \includegraphics[width=0.48\textwidth]{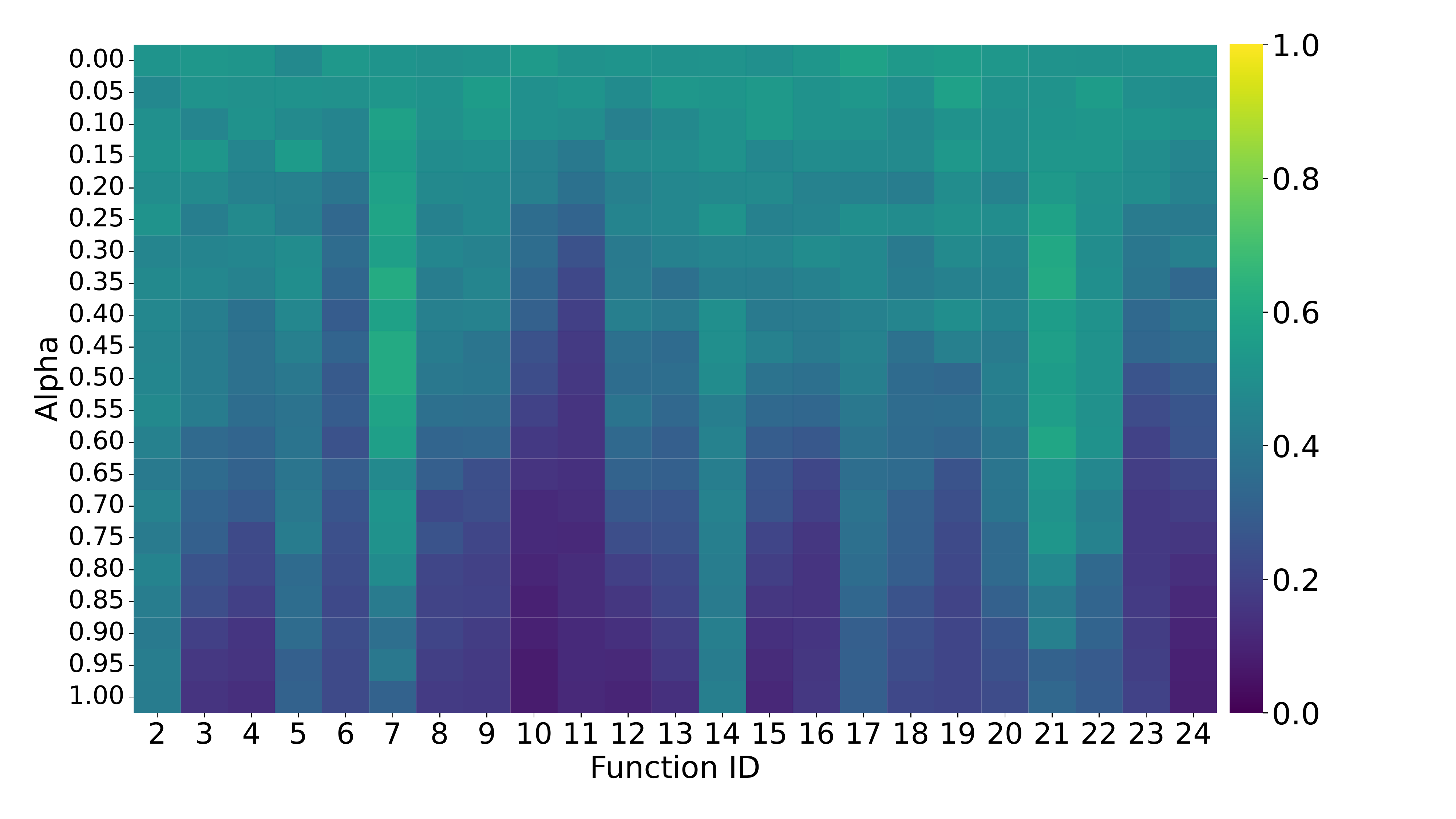}
    \caption{Normalized area under the ECDF curve of Differential Evolution for each combination of the BBOB-function (x-axis) with a sphere model, for given value of $\alpha$ (y-axis). AUC is calculated after $10\,000$ function evaluations, based on 50 runs on 10 instances.}
    \label{fig:de_heatmap_wsphere}
\end{figure}

\begin{figure}
    \centering
    \includegraphics[width=0.48\textwidth]{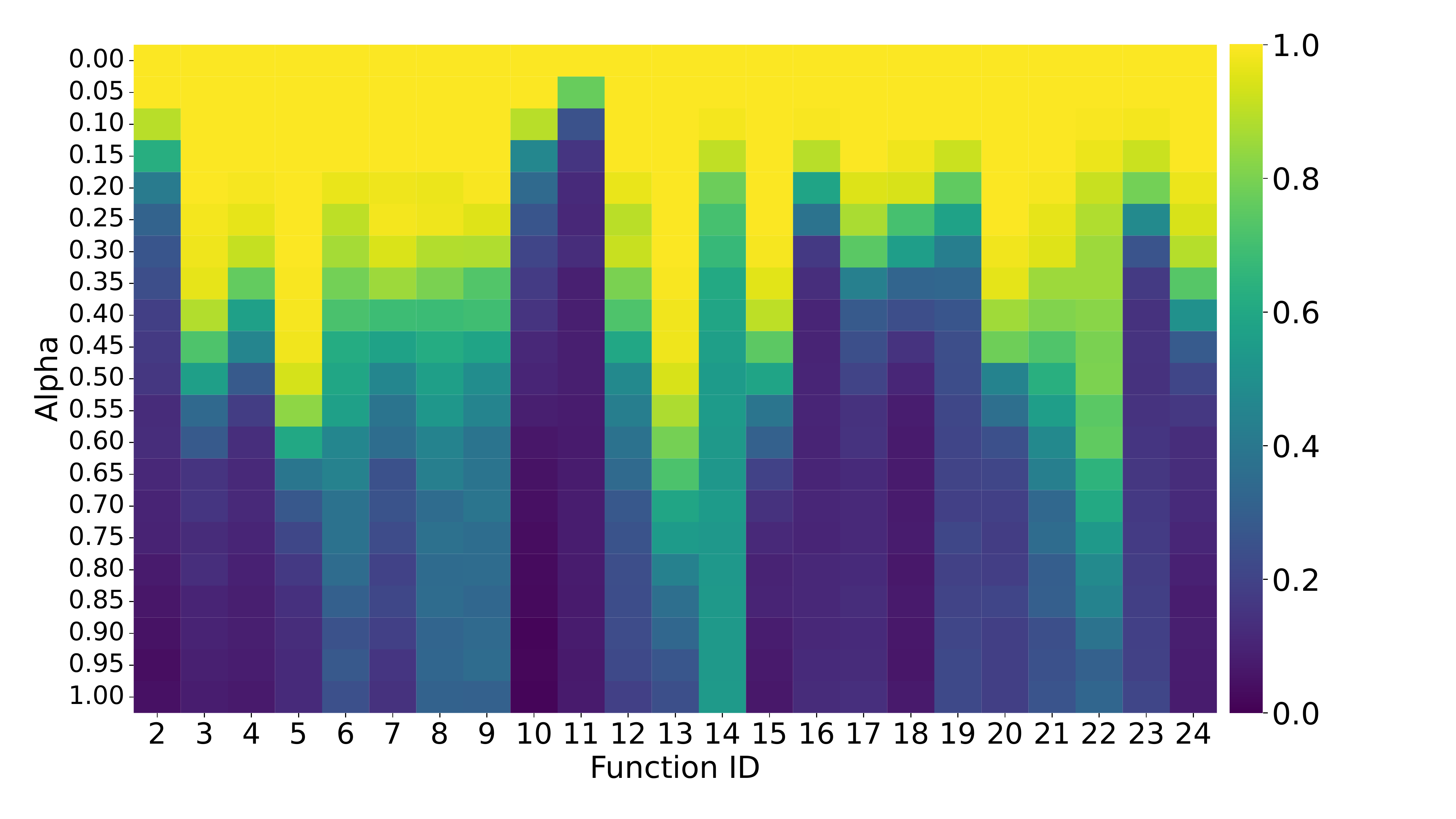}
    \caption{Normalized area under the ECDF curve of Cobyla for each combination of the BBOB-function (x-axis) with a sphere model, for given value of $\alpha$ (y-axis). AUC is calculated after $10\,000$ function evaluations, based on 50 runs on 10 instances.}
    \label{fig:cobyla_heatmap_wsphere}
\end{figure}

We can perform a similar analysis on other optimization algorithms. In Figure~\ref{fig:de_heatmap_wsphere} and Figure~\ref{fig:cobyla_heatmap_wsphere}, we show the same heatmap as Figure~\ref{fig:cma_heatmap_wsphere}, but for Differential Evolution and Cobyla respectively. It is clear from these heatmaps that the performance of DE is more variable than that of CMA-ES, while Cobyla's performance drops off much more quickly. The overall trendlines for DE do seem to be somewhat similar to those seen for diagonal CMA-ES: the transition points between high and low AUC in Figure~\ref{fig:de_heatmap_wsphere} are comparable to those seen in Figure~\ref{fig:cma_heatmap_wsphere}. There are however still some differences in behavior, espcially relative to Cobyla. These differences then lead to the question of whether there exist transition points in ranking between algorithms as well. Specifically, if one algorithm performs well for $\alpha=0$ but gets overtaken as $\alpha\rightarrow1$, exploring this change in ranking would give further insight into the relative strengths and weaknesses of the considered algorithms. 

\begin{figure}
    \centering
    \includegraphics[width=0.48\textwidth]{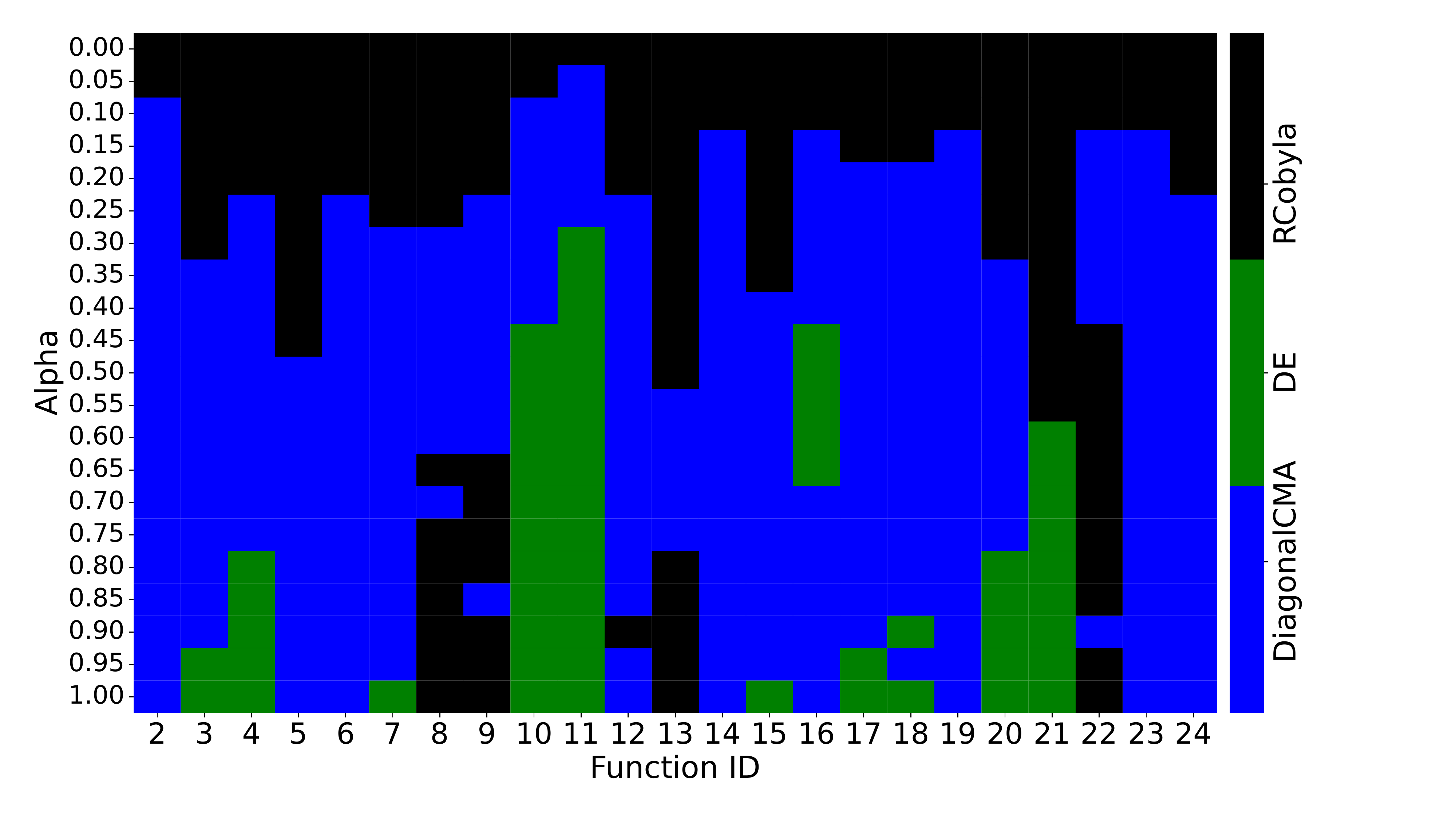}
    \caption{Algorithm with the highest area under the ECDF-curve for each combination of the BBOB-function (x-axis) with a sphere model, for given value of $\alpha$ (y-axis). AUC is calculated after $10\,000$ function evaluations, based on 50 runs on 10 instances. PSO and EMNA are not shown since they never ranked first.}
    \label{fig:rank_grid}
\end{figure}

In order to answer this question about the relative ranking of algorithms, we make use of the portfolio of 5 algorithms and rank them based on AUC on each affine function combination. We then visualize the top ranking algorithm on each setting in Figure~\ref{fig:rank_grid}. Important to note is that both PSO and EMNA never ranked first for the selected budget, and are thus not visible on the figure. 

From Figure~\ref{fig:rank_grid}, we can clearly see that Cobyla deals well with the sphere model, managing to outperform the other algorithm when the weighting of the sphere is relatively high. Then, after a certain threshold, the CMA-ES consistently outperforms the rest of the portfolio. However, as $\alpha$ increases further, and the influence of the sphere model diminishes, an interesting pattern seems to occur. For several problems, there is a second transition point, to either DE or Cobyla. For some functions, e.g. F3 and F4, one factor which might explain this phenomenon is the strength of the local optima increasing, making it harder for CMA-ES to explore the full landscape, while the uniform initialization of DE causes it to be slightly less impacted.

\begin{figure*}
    \centering
    \includegraphics[width=0.97\textwidth]{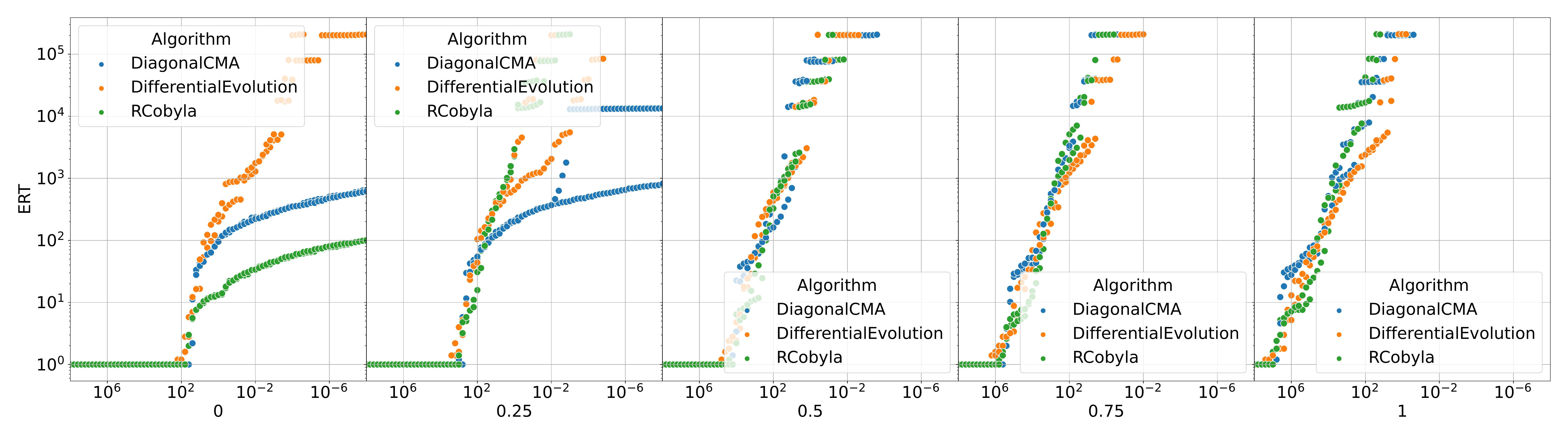}
    \caption{ERT per instance for three algorithms on the affine combinations between F10 
    ($\alpha =1 $) and F1 ($\alpha=0$), for selected values of $\alpha$. Each dot corresponds to the ERT calculated based on 5 runs on 1 instance, for a total of 10 instances. }
    \label{fig:ert_grid}
\end{figure*}

In order to better understand what the transitions in algorithm ranking look like, we can zoom in on one of the functions and plot the expected running time (ERT) for several values of $\alpha$. This is done in Figure~\ref{fig:ert_grid}, where we look at the combination between F10 and the sphere model. we clearly see that Cobyla is very effective at optimizing the sphere model, solving it almost an order of magnitude faster than the second ranked algorithm, which is DiagonalCMA. However, when $\alpha$ increases, Cobyla quickly starts to fail, while DiagonalCMA still manages to solve most instances at  $\alpha=0.25$ within similar amounts of evaluations. However, it is clear from the bifurcation in the plot that on some instances, the DiagonalCMA is no longer able to find the optimum within the allocated budget. When $\alpha$ increases further, none of the instances are able to be solved anymore by any of the three algorithms. When $\alpha\geq0.75$, we see that DE overtakes the other two, which explains the better ranking seen in Figure~\ref{fig:rank_grid}. 

\section{Combinations between Different Function Groups}

\begin{figure}
    \centering
    \includegraphics[width=0.48\textwidth]{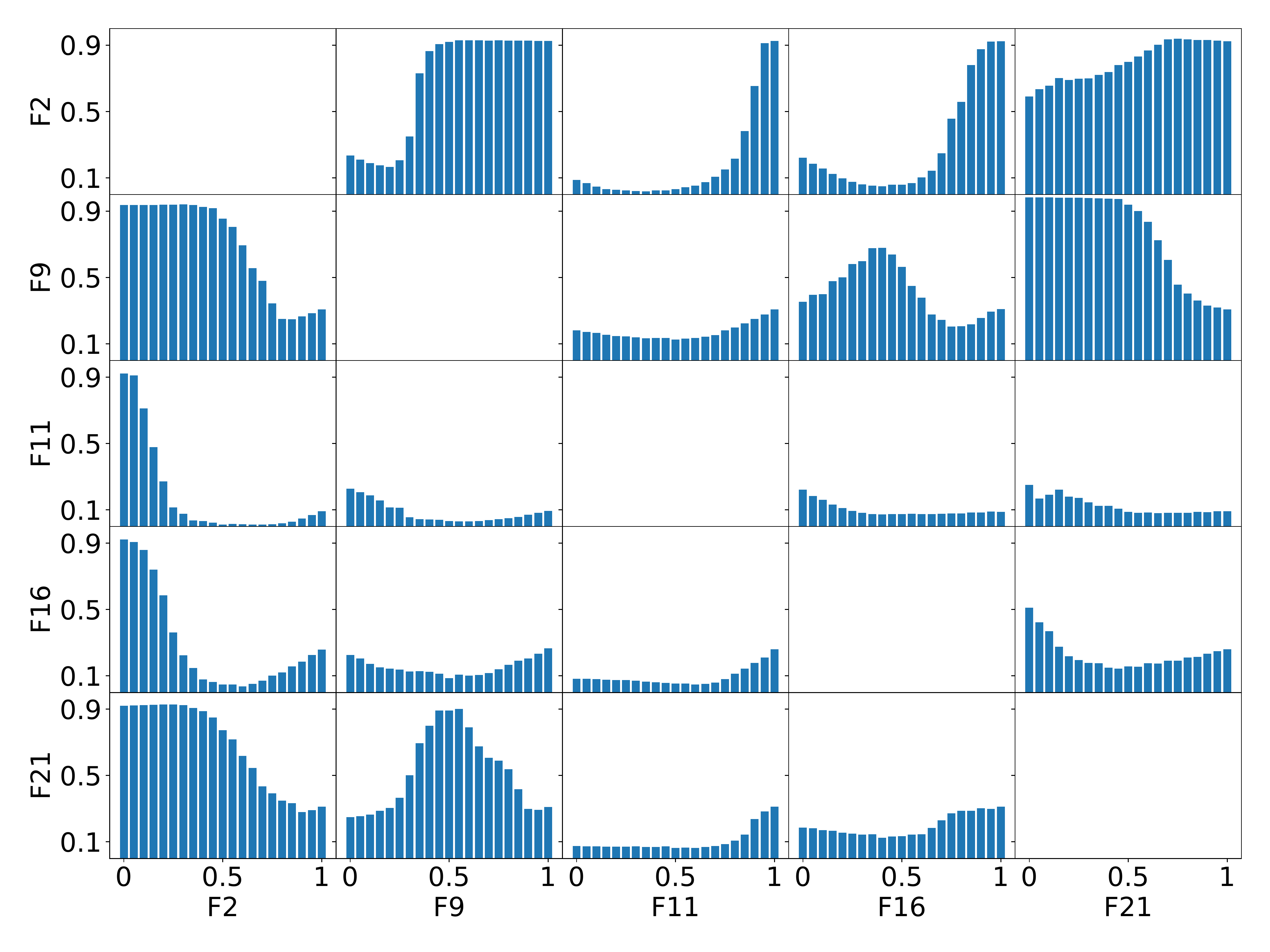}
    \caption{Area under the ECDF-curve for Diagonal CMA-ES on each of the affine combinations between the selected BBOB problems. Each facet corresponds to the combination of the row and column function, with the x-axis indicating the used $\alpha$. AUC values are calculated based on 50 runs on 5 instances, with a budget of $10\,000$ function evaluations. }
    \label{fig:cma_grid}
\end{figure}

While combining functions with a sphere model can be viewed as adding global structure to a problem, combinations between other functions can provide interesting insights into the transition points between different types of problems. To illustrate the kinds of insights that can be gained from these combinations, we select a subset of 5 functions and collect performance data on each combination with the same 21 $\alpha$ values (with both orderings of the function). We show the performance in terms of normalized AUC of diagonal CMA-ES on these function combinations in Figure~\ref{fig:cma_grid}. Note that for $\alpha=1$, we are using the function specified in the column label, while for $\alpha=0$ we have the function specified in the row label, but with the optimum of the column function. 

From Figure~\ref{fig:cma_grid}, we can see that the transition of performance between the two extreme $\alpha$ values is mostly smooth. While there are some rather quick changes, e.g. for the transition between F2 and F11, these seem to be the exception rather than the rule. Particularly interesting are the settings where the performance of affine combinations between two functions proves to be much easier or harder than the functions which are being combined. This is the case e.g. for the combinations of F21 and F9. 
Of note in this function combination is the fact that its mirrored combination around the diagonal does not display similar behavior. In fact, Figure~\ref{fig:cma_grid} in general is not fully symmetric around the diagonal. 

We might expect $(F_1, F_2, \alpha)$ to be similar to $(F_2, F_1, 1-\alpha)$. However, the combination between F9 and F21 shows that this is not always the case. Specifically, the AUC for the combination $(F_{21}, F_{9}, 1)$ is significantly worse than that of $(F_{9}, F_{21}, 0)$, even though F21 does not contribute directly to the function value of the affine combination. The only way in which these two problems differ is in the location of the optima. For F21, the default location of the optimum is hard-coded to be at distance 1 from the optimum~\cite{evostar_bbob_instance}, which is not the case for F9. Since the CMA-ES initializes its center of mass in the origin of the space and uses a default initial stepsize of $0.3$~\cite{nevergrad}, it is able to find the optimum in the default setting, while the translated version of the function becomes much more challenging. This highlights a potential issue with the traditional analysis of performance on BBOB problems: if we don't take into account the built-in limitations on e.g. the location of the optimum in our analysis, there is a risk of misinterpreting the results of a structurally biased algorithm~\cite{bias} and viewing it as optimal on this type of multimodal problem, while it is unable to solve a translated version of the same function.

\begin{figure}
    \centering
    \includegraphics[width=0.48\textwidth]{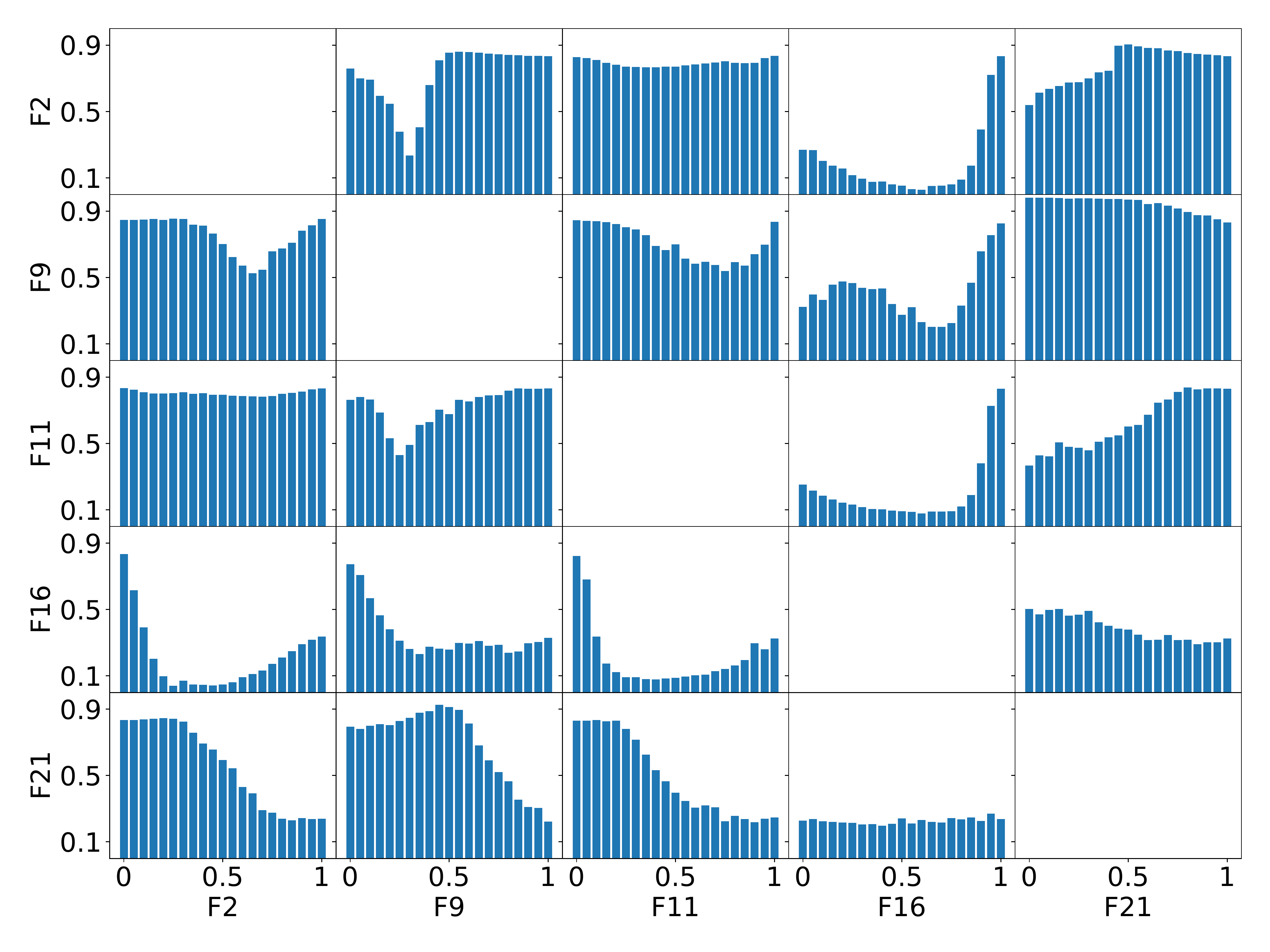}
    \caption{Area under the ECDF-curve for modular CMA-ES on each of the affine combinations between the selected BBOB problems. Each facet corresponds to the combination of the row and column function, with the x-axis indicating the used $\alpha$. AUC values are calculated based on 50 runs on 5 instances, with a budget of $10\,000$ function evaluations.}
    \label{fig:modcma_grid}
\end{figure}

To see how much this initialization really impacts the differences in performance, we perform an additional experiment with a different version of CMA-ES. We opt to use the modular CMA-ES~\cite{nobel_modcma_assessing} and set the initial stepsize to $0.2$ times the range of the domain, so $2$ in our case. The resulting performance is visualized in Figure~\ref{fig:modcma_grid}. In this figure, it is clear that the overall performance of this setting of CMA-ES performs better overall, but of particular note is that the asymmetries have been somewhat reduced, although not disappeared entirely. 

Additionally, Figure~\ref{fig:modcma_grid} shows several interesting trends in performance which were not present for the Diagonal CMA-ES. For example, the combinations between F2 and F9 show a large dip in AUC near the center, even though both functions separately seem relatively easy to solve for this version of CMA-ES. While the differences between the two versions of CMA-ES are noticeable, many of the trends, e.g. decreased performance for combinations between F11 and F16, are present to some extent in Figure~\ref{fig:cma_grid} as well. 

\begin{figure}
    \centering
    \includegraphics[width=0.48\textwidth]{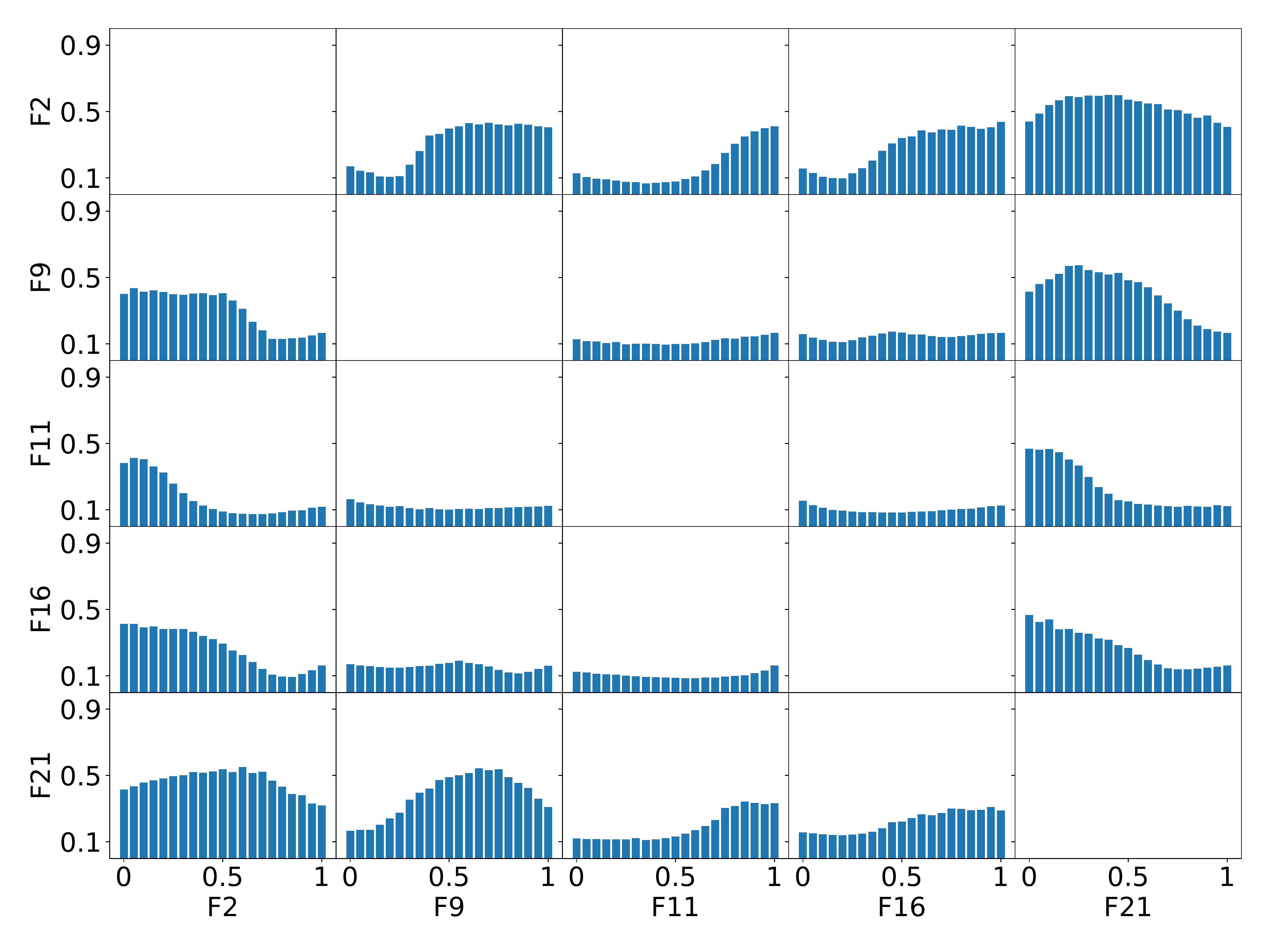}
    \caption{Area under the ECDF-curve for Differential Evolution on each of the affine combinations between the selected BBOB problems. Each facet corresponds to the combination of the row and column function, with the x-axis indicating the used $\alpha$. AUC values are calculated based on 50 runs on 5 instances, with a budget of $10\,000$ function evaluations.}
    \label{fig:de_grid}
\end{figure}

As a final algorithm, we run DE on the same set of function combinations. The results are visualized in Figure~\ref{fig:de_grid}. In this figure, we see that the overall performance of DE is indeed worse than the two versions of CMA-ES. It is worth noting that the amount of asymmetry along the diagonal is lower than for the diagonal CMA-ES. This could be caused by the change in initialization (Gaussian for CMA-ES, uniform for DE) reducing the initial bias to the center of the space. Another factor to consider is the variance of the performance. For CMA-ES, performance can vary significantly as $\alpha$ changes, while the changes in AUC seem to be much smaller for DE.

\begin{figure}
    \centering
    \includegraphics[width=0.48\textwidth]{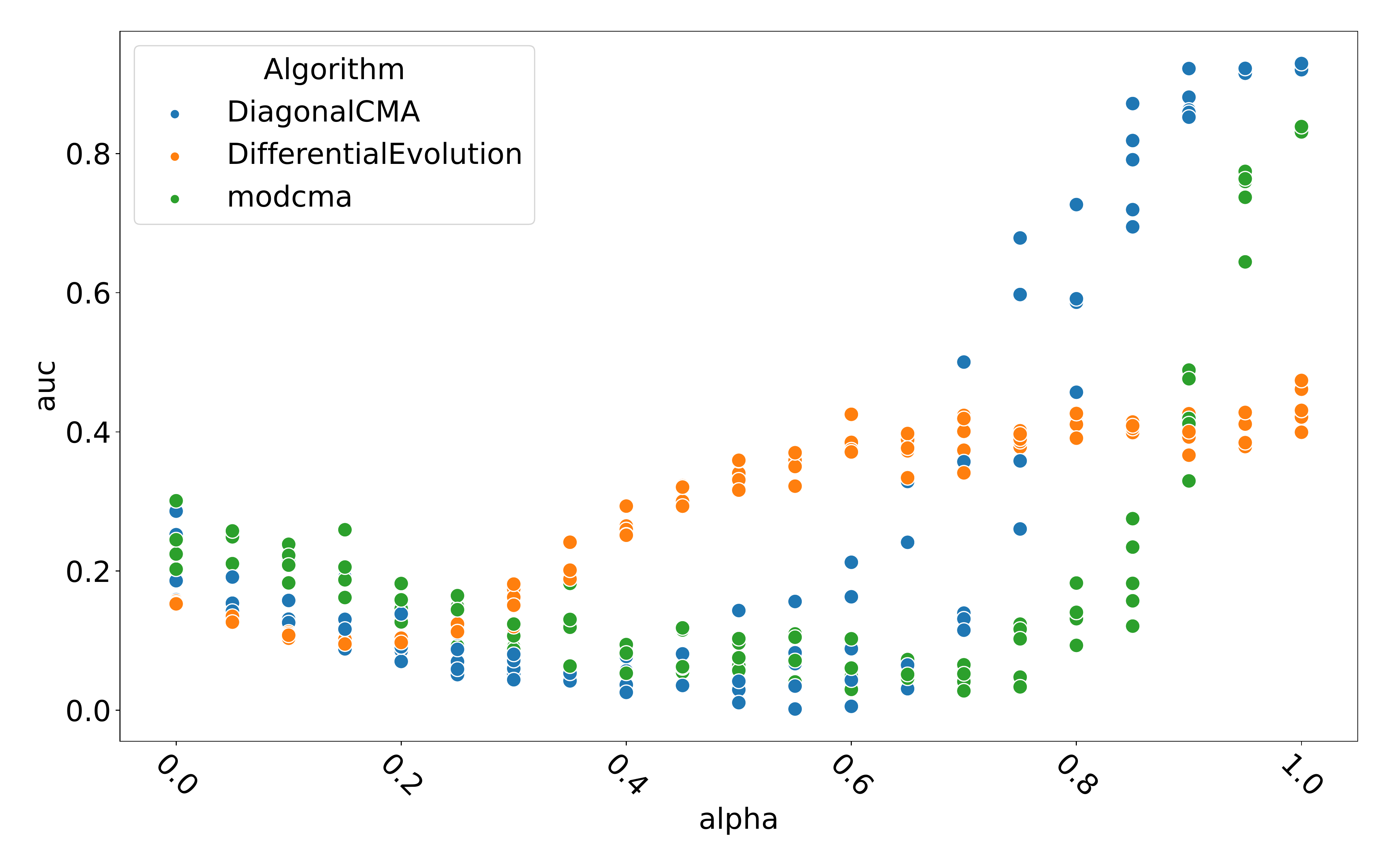}
    \caption{Distribution of per-instance normalized AUC values for the selected algorithm on the affine combination between F2 and F16. AUC values are calculated based on 50 runs on 5 instances, with a budget of $10\,000$ function evaluations.}
    \label{fig:f2_to16_scatter}
\end{figure}

\begin{figure}
    \centering
    \includegraphics[width=0.48\textwidth]{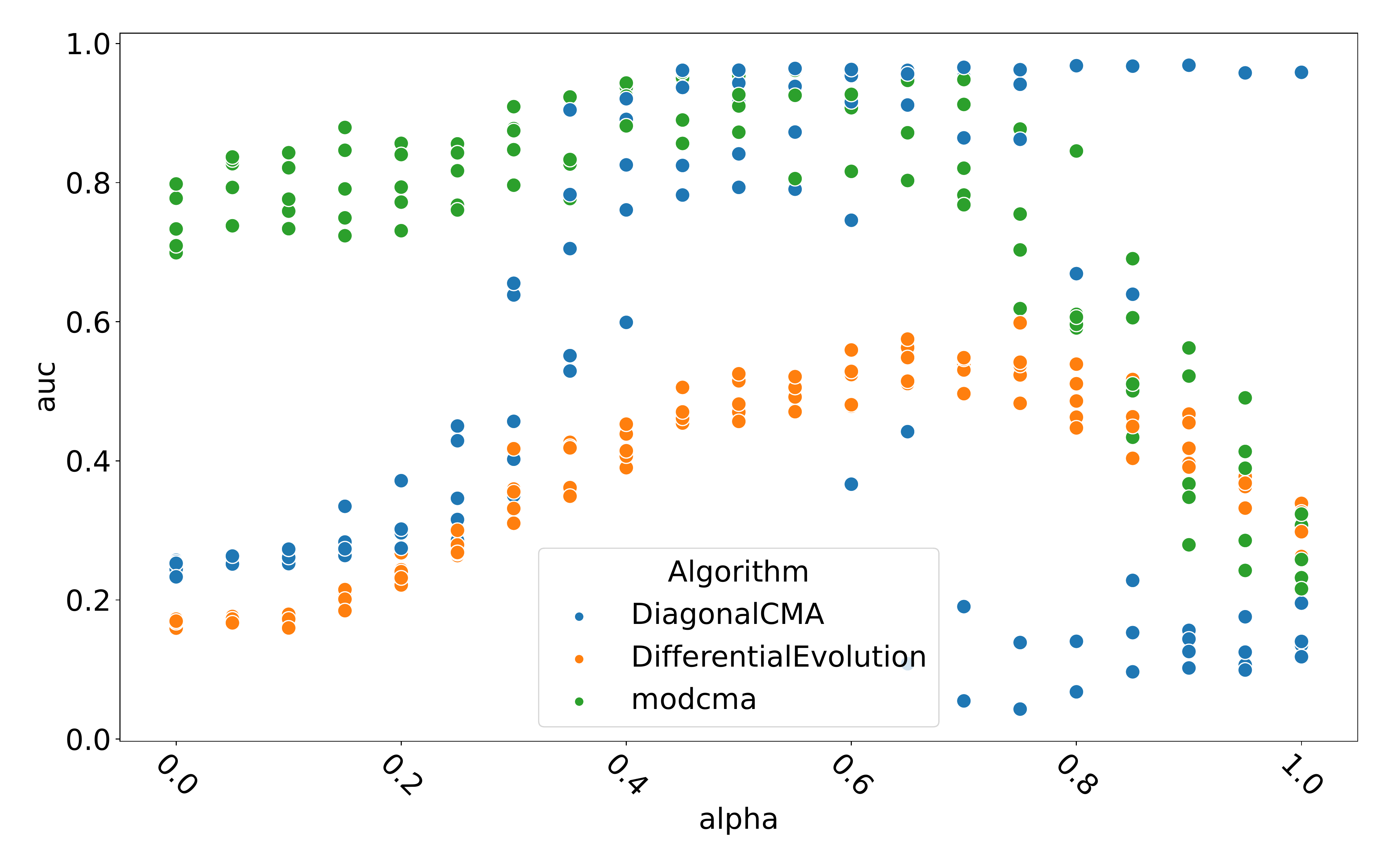}
    \caption{Distribution of per-instance normalized AUC values for the selected algorithm on the affine combination between F21 and F9. AUC values are calculated based on 50 runs on 5 instances, with a budget of $10\,000$ function evaluations.}
    \label{fig:f21_to9_scatter}
\end{figure}

\section{Zooming Into One Function Combination}\label{sec:zooming}

To further analyze the impact of changing the weighting of the function combinations, we can zoom in on one particular combination and study it in more detail. First, we gauge the impact of using different instances to measure performance. This is done by considering the distribution of AUC values for a specific function combination, F2 to F16, in Figure~\ref{fig:f2_to16_scatter}, on a per-instance basis. From this figure, we see that in general, the distribution of AUC values is rather stable. However, at the transition point for the CMA-ES variants, around $\alpha\approx 0.8$, we see a clear increase in variance. To check whether this behavior also occurs for other function combinations, we create the same visualization for the combination of F21 and F9 in Figure~\ref{fig:f21_to9_scatter}. In this figure, we see a similar pattern for the diagonal CMA-ES, where the distribution of AUC at high $\alpha$ ranges from almost $0$ to almost $1$. 

The variance observed in Figure~\ref{fig:f2_to16_scatter} 
might indicate that, in order to get a stable view of the exact behavior at this transition point, a wider variety of instances should be used to get a more robust performance estimate. However, when considering the extreme differences in AUC observed in Figure~\ref{fig:f21_to9_scatter}, this variance invites a more detailed study into the interaction between the instance generation process (e.g., the placement of the optimal solution) and the search behavior of the used algorithm.  

\begin{figure}
    \centering
    \includegraphics[width=0.48\textwidth]{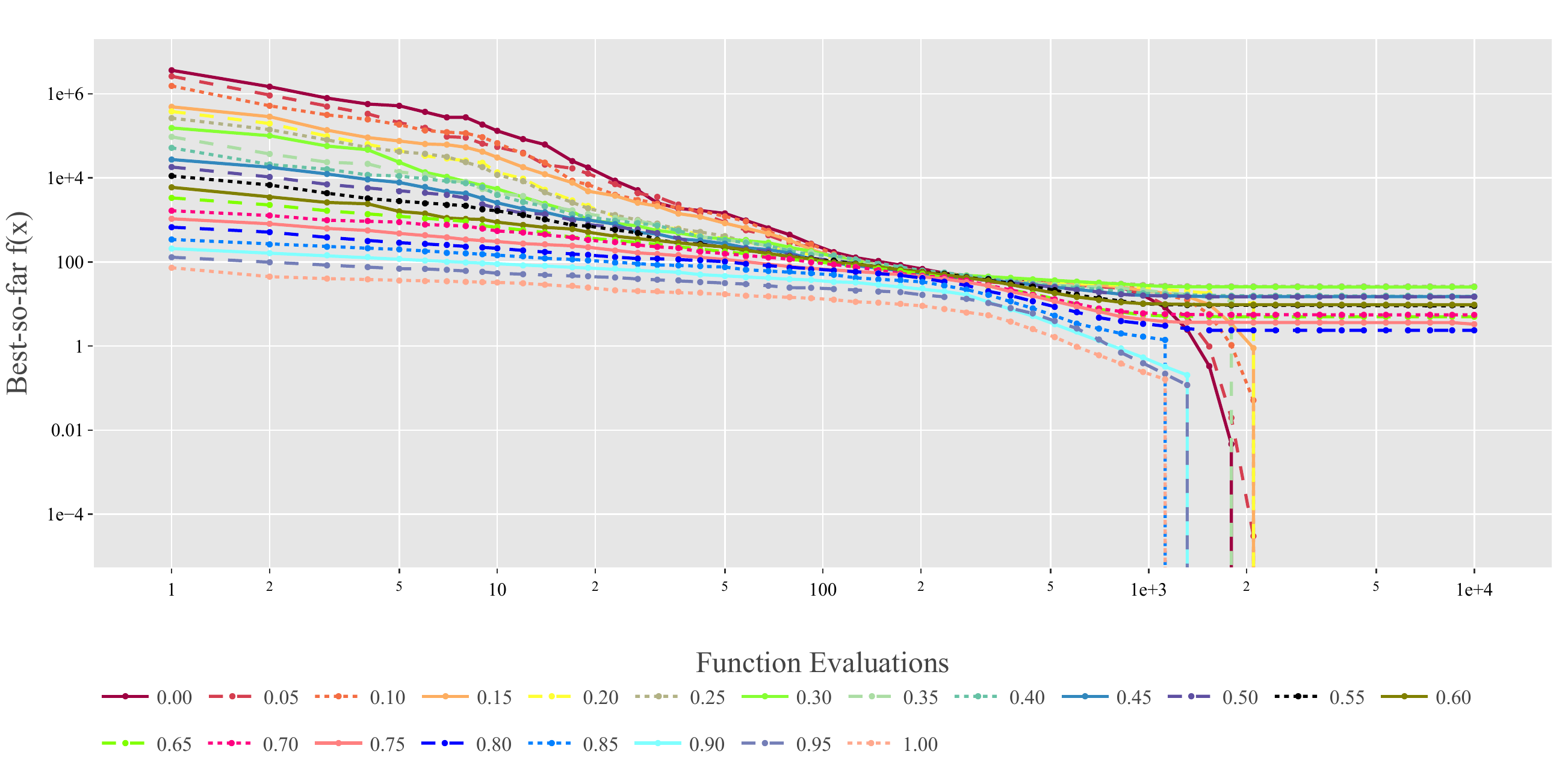}
    \caption{Evolution of geometric mean function value found by modular CMA-ES for the affine combination of F16 and F11, instance 1 for both functions. Each line corresponds to 50 runs with the specified $\alpha$. }
    \label{fig:f16_to11_covergence_modcma}
\end{figure}

Next to the instance generation process, another important factor to consider when analyzing the performance of optimization algorithms on these affine function combinations is the scaling of the objective values. While it is common practice to ignore the scaling, so the same target values (precision to the optimum) can be used, for example to compute aggregated ECDF curves, the ways in which different problems scale their objective values does influence how we should interpret their results. This becomes increasingly obvious when considering the affine combinations of these problems. In Figure~\ref{fig:f16_to11_covergence_modcma}, we show the convergence plot of diagonal CMA-ES on the combination of F16 and F11. We clearly see from the left part of this curve that the initial values found vary widely for different combinations, ranging from $10^{7}$ when $\alpha=0$ to $10^2$ when $\alpha=1$. However, the change in scale is not the only factor impacting the performance. The shape of the curve changes noticeably after the initialization, which matches the change in AUC observed in Figure~\ref{fig:cma_grid}. 

\begin{figure*}
    \centering
    \includegraphics[width=0.97\textwidth]{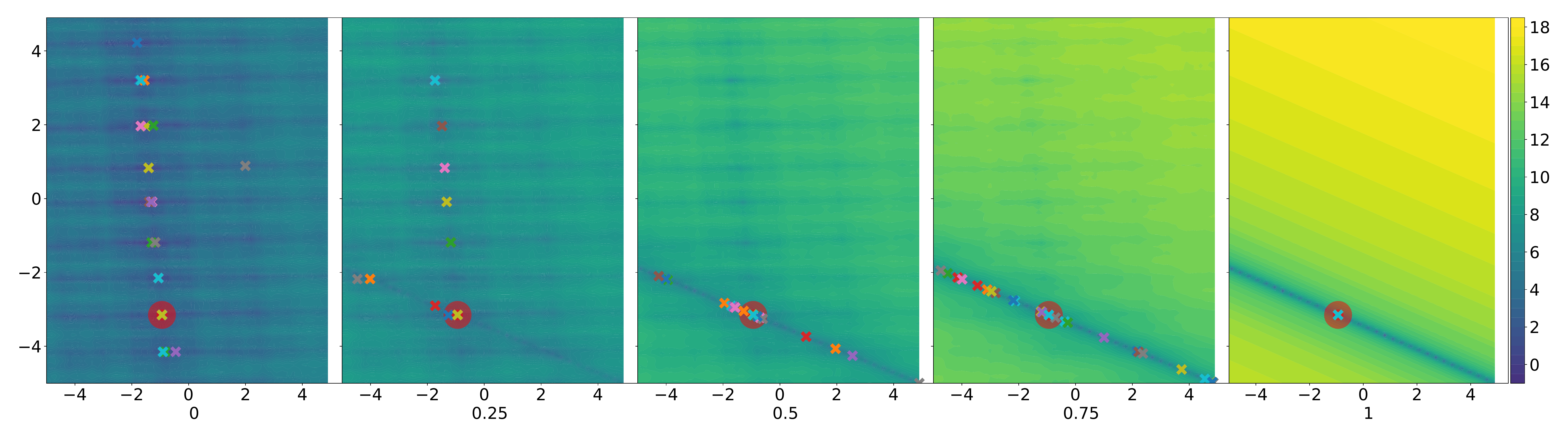}
    \caption{Evolution of the landscape (log-scaled function-values) of the affine combination between F11 ($\alpha=1$) and F16 ($\alpha=0$), instance 1 for both functions, for varying $\alpha$. The red circle highlights the location of the global optimum. The crosses correspond to the best point found in each of 50 runs of the modular CMA-ES.}
    \label{fig:frame_f11_16}
\end{figure*}

\begin{figure*}
    \centering
    \includegraphics[width=0.97\textwidth]{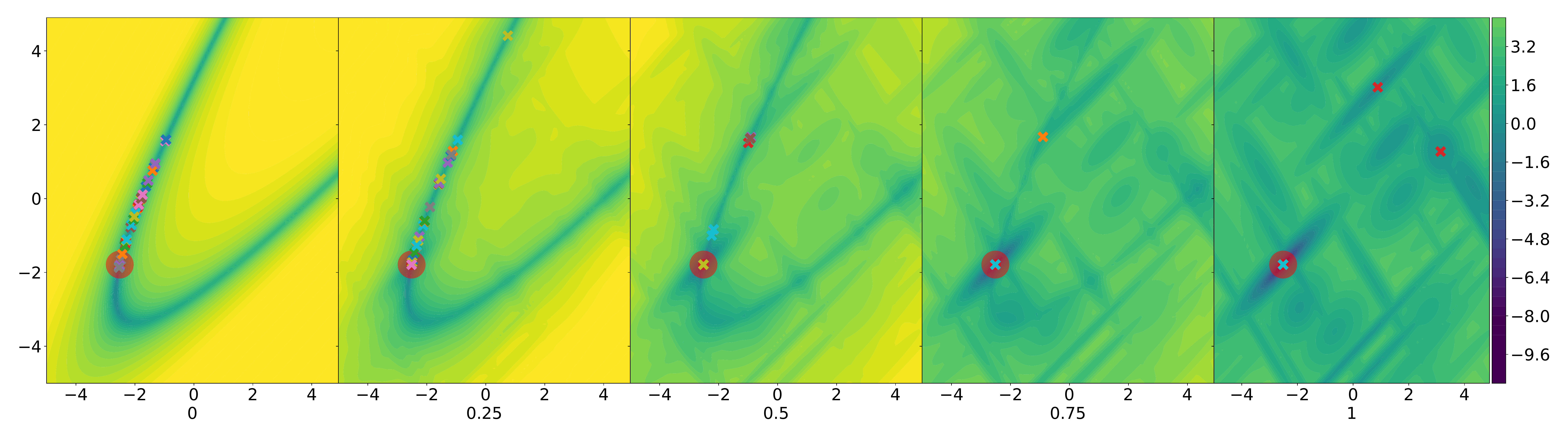}
    \caption{Evolution of the landscape (log-scaled function-values) of the affine combination between F21 ($\alpha=1$) and F9 ($\alpha=0$), instance 1 for both functions, for varying $\alpha$. The red circle highlights the location of the global optimum. The crosses correspond to the best point found in each of 50 runs of Diagonal CMA-ES.}
    \label{fig:frame_f21_9}
\end{figure*}

To investigate the reason for this change in behavior, we can study the optimization trajectory of diagonal CMA-ES on these functions. Since this is not feasible to visualize in the original 5-dimensional space, we repeat the data collection on the 2-dimensional version of these functions. In Figure~\ref{fig:frame_f11_16}, we show the landscapes of the affine combinations between F11 and F16 for several values of $\alpha$. We highlight the best point found by the diagonal CMA-ES in each of its 50 runs on this instance. This plot clearly shows the differences in scale between the original problems. In addition, we see that as $\alpha$ gets closer to 1, the algorithm gets stuck in the local optima less often. The global structure added by F11 is strong enough to guide the CMA-ES to the area containing the global optimum. However, when the influence of F11 becomes too large, the difficulties of finding the correct search direction have a strong impact on the convergence behavior. As such, values of $\alpha$ closer to $0.5$ seem to provide a mix of the multimodality of F16 and the challenges of F11, which makes it a challenging problem to solve for the CMA-ES. 

While the combination between F11 and F16 seems to create functions that are more challenging, Figure~\ref{fig:cma_grid} shows that there are function combinations where the opposite is true. The combination between F9 and F21 displays interesting behavior. While the way of performing initialization might explain the asymmetry between $(F_{9}, F_{21}, 0)$ and $(F_{21}, F_{9}, 1)$, it does not explain the increase in AUC for $\alpha$ close to $0.5$. We visualize the change in landscape, and corresponding solutions found by the diagonal CMA-ES, in Figure~\ref{fig:frame_f21_9}. In this figure, we see that when $\alpha=0$, the CMA-ES finds solutions on the ridge of the function, but most of the runs don't reach the optimum within the given budget. This indicates that the characteristic difficulty of F9, the algorithm having to consistently adapt its search direction~\cite{bbobfunctions}, hinders the convergence of the used diagonal CMA-ES.
However, as $\alpha$ increases, the structure of F21 gets added, which increases the ways in which the algorithm can approach the optimum value. For $\alpha=1$, the multimodality from F21 completely takes over, trapping some runs in local optima, thus decreasing the performance of the algorithm. This showcases that combining these two functions in this way creates a function where the original difficulties of both are combined in a way that negates both of them, which is then exploited by the CMA-ES.

\section{Conclusions and Future Work}

Affine combinations of BBOB problems offer a new way to investigate the behavior of optimization algorithms. We have shown how combinations of arbitrary functions with a sphere model can be used to identify the impact of added global structure on the performance of a set of algorithms. In addition, combinations between functions with different high-level characteristics allowed us to observe transitions between different optimization challenges. While this investigation is not exhaustive, it highlights the potential benefit of utilizing these new function combinations for gaining an understanding of the behavior of optimization algorithms. 

However, these benefits in terms of analysis options also come with several challenges which have to be considered. We identified the following aspects:

\textbf{Scaling.} As identified when these combinations were proposed~\cite{affinebbob}, the differences in scale between two problems can be significant. While we aimed to reduce this impact by considering a logarithmically scaled weighting, it is clear from our experiments that the scale still plays a large role in the way we interpret the performance. Finding ways to combine the landscapes of two functions while maintaining a consistent range of function values is still an open question. 

\textbf{Instances.} The BBOB suite is built on the idea that each function can be instantiated in many ways. This is achieved through several transformations, the most common of which is moving the optimum to a different location in the domain. The results we present show that the way in which these optimal locations are chosen can have a large impact on the performance of optimization algorithms. Since the optima are not distributed uniformly in the domain, some functions have different kinds of bias, which can be exploited by an algorithm. The question on how to fairly consider different instance generation mechanisms when making use of function combination is thus highly interlinked with questions about how well performance observed on a set of BBOB instances generalizes. 

Even with these challenges in mind, there are many potential use cases for these affine function combinations. One aspect in which they can prove useful is in the training of algorithm selection models~\cite{kerschke2018survey}, as they can significantly increase the size and variety of training data, which is an important consideration towards testing generalizability.

One final aspect in which the benchmark data on these function combinations can be further utilized is by linking it back to the exploratory landscape analysis which inspired their creation. Since the combinations can smoothly fill the landscape feature space, this can be combined with algorithm performance to get a more fine-grained view of the way in which the landscape interacts with different algorithms~\cite{trajanov2021explainable, jankovic2020landscape}.

\begin{acks}
Our work is financially supported by ANR-22-ERCS-0003-01 project VARIATION and by the CNRS INS2I project IOHprofiler. This work was performed using the ALICE compute resources provided by Leiden University. 
\end{acks}

\bibliographystyle{ACM-Reference-Format}
\bibliography{bib/abbrev,bib/journals,bib/authors,bib/biblio,bib/crossref,references}

\end{document}